\PassOptionsToPackage{table}{xcolor}

\documentclass[preprint,12pt,authoryear]{elsarticle}
\usepackage[table]{xcolor}

\usepackage{amsmath,amsfonts,bm}

\def\eqref#1{equation~\ref{#1}}

\def\1{\bm{1}}

\DeclareMathAlphabet{\mathsfit}{\encodingdefault}{\sfdefault}{m}{sl}
\SetMathAlphabet{\mathsfit}{bold}{\encodingdefault}{\sfdefault}{bx}{n}

\usepackage{natbib}
\setcitestyle{numbers,square}
\usepackage{hyperref}
\usepackage{url}
\usepackage{algorithm}
\usepackage[noend]{algpseudocode}
\usepackage{algorithmicx}

\usepackage{multirow}  
\usepackage{booktabs}  
\usepackage{array}     
\usepackage{multirow, booktabs, array, tabularx}
\usepackage{ragged2e}
\usepackage{multirow, booktabs, array, tabulary}
\usepackage{ragged2e}
\usepackage{subcaption}   

\usepackage{graphicx}

\usepackage{wrapfig}   

\journal{Neural Networks}

\begin{document}

\begin{frontmatter}
 \author{Senmao Ye\textsuperscript{a}, Fei Liu\textsuperscript{b}, Dawei Zhang\textsuperscript{a},Yuqian Zheng\textsuperscript{c},Jun Zhang\textsuperscript{a}, Zhonglong zheng\textsuperscript{a*}}
 
 \affiliation[label1]{organization={Zhejiang Normal University},
 	city={Jinhua},
 	postcode={321004}, 
 	state={Zhejiang},
 	country={China}}
 \affiliation[label2]{organization={ South China University of Technology},
 	city={Guangzhou},
 	postcode={510006}, 
 	state={Guangdong},
 	country={China}}
 	 \affiliation[label3]{organization={ Jinhua University of Vocational Technology},
 		city={Jinhua},
 		postcode={321007}, 
 		state={Zhejiang},
 		country={China}}
 \tnotetext[label1]{Corresponding author.\\ \indent Email address:  senmaoy@gmail.com,  feiliu@scut.edu.cn,davidzhang@zjnu.edu.cn,\\everygreen870@gmail.com,junzhang@ieee.org,zhonglong@zjnu.edu.cn}
 \title{Regularized Wasserstein Distance for Discrete Generative Modeling and Clustering Analysis} 
\begin{abstract}
	In this paper, we formulate clustering analysis as discrete generative modeling. Specifically, when the generator is restricted to a finite candidate space, the generated samples naturally emerge as the most representative prototypes of the underlying data distribution. To model such a discrete candidate space, we employ a Discrete Distribution Network (DDN). However, DDN suffers from unstable training due to its hard nearest-neighbor selection. To stabilize training and encourage geometric consistency, we replace DDN's $\ell_2$ loss between images with regularized Wasserstein distance between the empirical distributions of real and generated images. The resulting framework, termed Discrete Generative Clustering (DGC), offers four notable advantages: 1) the learned cluster centers are directly interpretable, as they are realistic synthetic samples; 2) the inference cost scales logarithmically with the number of clusters, i.e., $O(\log K)$, making the method scalable to extremely large $K$; 3) the regularized Wasserstein distance provides a smooth and well-behaved optimization objective that facilitates stable convergence, whereas DDN's $\ell_2$ loss tends to converge to poor local optima; and 4) DGC replaces DDN's nearest-neighbor selection with random-neighbor selection, which avoids the density shift problem during multi-step forward. Extensive experiments on several benchmarks demonstrate the effectiveness and desirable properties of DGC. The code is publicly available at \href{https://github.com/senmaoy/generative-discrete-clustering.git}{GitHub}.
\end{abstract}


\end{frontmatter}



\section{Introduction}
Clustering analysis is a fundamental task in unsupervised learning, aiming to partition data into meaningful groups without label supervision~\cite{everitt2011cluster,delbarrio2019robust,mukherjee2019clustergan,ester1996dbscan,JIANG2027109382,PENG2027109372,QIU2026108973}. It plays a critical role in image segmentation, bioinformatics, anomaly detection, and exploratory data analysis. Despite decades of research, traditional clustering methods still suffer from several inherent limitations. Partition-based algorithms like K-Means~\cite{lloyd1982least} require a predefined number of clusters $K$ and are highly sensitive to initialization. Model-based approaches such as Gaussian Mixture Models (GMM)~\cite{bishop2006pattern} assume specific parametric forms and become computationally expensive as $K$ grows. Hierarchical methods incur $\mathcal{O}(n^2)$ complexity~\cite{mullner2011modern}, and density-based methods like DBSCAN~\cite{ester1996density,liu2026efficient} are sensitive to hyperparameter choices. Moreover, many of these methods yield cluster centers that are abstract mathematical constructs, lacking direct physical interpretability~\cite{guyon2012clustering}.

In recent years, generative models~\cite{ho2020denoising, DBLP:conf/nips/SongE19,
	DBLP:conf/iclr/0011SKKEP21, hyvarinen2005estimation} have demonstrated remarkable
capability in modeling complex data distributions, providing new paradigms for tasks
such as image segmentation, image classification, and object detection. For clustering
analysis, existing models~\cite{mukherjee2019clustergan, yan2025clusterddpm,
	DBLP:conf/ijcai/JiangZTTZ17} invert the generative process to extract unsupervised
deep feature representations. However, the clustering step is still accomplished by
conventional methods such as K-means and GMM. These generation-based cluster models
do not further explore the intrinsic relationship between clustering and generation.

\begin{wrapfigure}{r}{0.50\textwidth}
	\vspace{-0.5cm}
	\centering
	\includegraphics[width=\linewidth]{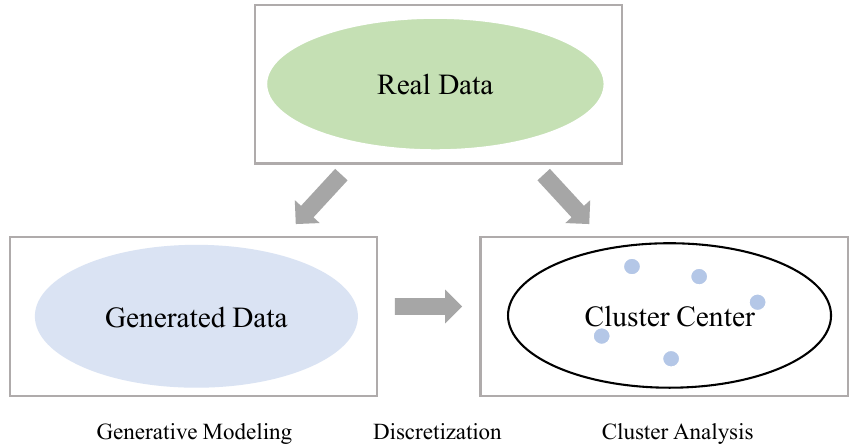}
	\caption{Clustering analysis is a special case of generative modeling. }
	\label{fig:gdc}
	\vspace{-0.7cm}
\end{wrapfigure}

To connect clustering analysis and generative modeling, this paper introduces Discrete Generative  Clustering (DGC). As depicted in Figure~\ref{fig:gdc}, the core idea is to generate a set of candidate samples that faithfully represent the entire dataset.  We first discretize generative modeling via a hard selection operation proposed by Discrete Distribution Network~\cite{yang2025discrete}. To preserve global geometric fidelity, we minimize the regularized Wasserstein distance~\cite{cuturi2013sinkhorn,qian2025incremental} between the distribution of generated candidates and the real data distribution. The regularized Wasserstein distance is differentiable and computationally efficient and better preserves the underlying manifold structure than DDN's image-wise loss. Finally, each data point is assigned to its nearest prototype according to the tree structure of DDN.

The proposed Discrete Generative Clustering offers several notable advantages. First, it provides interpretability, as the cluster centers are actual samples generated from the underlying data distribution, rather than abstract latent parameters. DGC provides a probabilistic interpretation of image clustering, grounded in generative modeling. Second, it exhibits high computational efficiency, with a complexity of $O(\log K)$, which remains tractable for extremely large numbers of clusters. Third, it ensures optimization stability: the regularized Wasserstein distance defines  a smooth and well-behaved optimization objective, in contrast to DDN's original $\ell_2$-based loss functions that are prone to falling into local optima. Fourth,  DGC is free of DDN's density shift problem caused by  multi-step  nearest-neighbor selection in training. Such training bias will accumulate when the number of DDN layers grow. Extensive experimental results demonstrate that DGC achieves competitive clustering performance while yielding visually interpretable cluster centers. Ablation studies further substantiate the individual contribution of each component. To facilitate reproducibility, the source code is made publicly available.
\section{Background}

\subsection{Regularized Wasserstein Distance}
Optimal transport (OT) provides a principled framework for comparing probability distributions by quantifying the minimal cost required to transform one distribution into another. Among OT metrics, the Wasserstein distance is particularly well-suited for generative modeling, as it preserves the geometric structure between distributions even when their supports are disjoint. However, computing the exact Wasserstein distance is computationally prohibitive for large-scale problems, since solving the associated linear program incurs a cost that is cubic in the sample size.

To mitigate this limitation, entropic regularization was introduced in \cite{cuturi2013sinkhorn}. Let $p$ and $q$ denote the empirical distributions of real samples and generated samples, respectively. The regularized Wasserstein distance is defined as
\begin{equation}
	W(p,q) = \min_{\pi\in\Pi(p,q)} \Bigl\{ \mathbb{E}_\pi[c(\mathbf{x},\mathbf{y})] + \lambda\, \mathrm{KL}(\pi \,\|\, p\otimes q) \Bigr\},\label{eq:sd_debiased}
\end{equation}
where $\lambda>0$ is the regularization parameter, and $c(\mathbf{x},\mathbf{y})$ is the ground-cost function representing the cost of transporting a unit mass from $\mathbf{x}$ to $\mathbf{y}$. The entropic term renders the optimization problem strictly convex, thereby enabling efficient computation via the Sinkhorn–Knopp matrix scaling algorithm. Consequently, the regularized Wasserstein distance is also referred to as the Sinkhorn–Knopp loss. Moreover, the Sinkhorn–Knopp iterations are fully differentiable, allowing end-to-end backpropagation through the transport plan. These properties make the regularized Wasserstein distance an ideal loss function for training deep generative models, providing a smooth, robust, and computationally tractable alternative to the exact Wasserstein distance. 
\subsection{Discrete Distribution Networks}
The DDN model consists of $L$ layers of Discrete Distribution Layers. For a given layer $l$, denoted as $G_l$, the input is the selected sample from the previous layer, $\mathbf{x}_{l-1}$. The layer generates $B$ new samples, $G_l(\mathbf{x}_{l-1})$, from which we select the sample $\mathbf{x}_l$ that is closest to the current training sample $\mathbf{x}$, along with its corresponding index $b_{l}$. The loss $J_l$ for this layer is then computed only on the selected sample $\mathbf{x}_l$.


\begin{align}
	b_{l} &= \underset{b \in \{1, \dots, B\}}{\operatorname{argmin}} \; \left\| G_l  (\mathbf{x}_{l-1})[b_{l-1}] - \mathbf{x} \right\|^2 \label{eq:k_star} \\
	\mathbf{x}_l &= G_l(\mathbf{x}_{l-1})[b_l] \label{eq:x_star} \\
	J_l &= \left\| \mathbf{x}_l - \mathbf{x} \right\|^2 \label{eq:J_l}
\end{align}

\section{Regularized Wasserstein Distance for Generative Modeling and Clustering}\label{sec:method}

To generate geometrically consistent prototypes, we first minimize the regularized Wasserstein distance between real and generated samples. Then, we discretize this generative model to enforce a discrete sample space. Finally, data points are assigned to the nearest prototypes according to the tree structure of DDN.
\subsection{Continuous  Generative Modeling with Regularized Wasserstein Distance}
%
Generative modeling is achieved by training a neural network generator \(G_\theta\) to minimize the regularized Wasserstein distance between the real and generated distributions. Let \(z \sim p(z)\) be a latent variable drawn from a fixed prior (e.g., a standard Gaussian), and let \(y = G_\theta(z)\) denote the generated sample. Given a mini-batch of real samples \(\textbf{x}=\{x_1,\dots,x_n\}\) and generated samples \(\textbf{y}=\{y_1,\dots,y_m\}\) with \(y_j = G_\theta(z_j)\), the generator is trained by minimizing the  regularized Wasserstein distance between the two empirical measures:
\begin{equation}
	\theta^* \;=\; \arg\min_{\theta} \; W\bigl(\mathbf{x}, \mathbf{y}),
	\label{eq:generative_sd}
\end{equation}
where \(W\) is defined as in \eqref{eq:sd_debiased}. With SGD training,  \( G_\theta(z)\) is able to synthesize reasonable images.
\subsection{Discrete  Generative Modeling  with Regularized Wasserstein Distance}

To enforce a discrete sample space, we discretize it using a Discrete Distribution Network.

\paragraph{Network Architecture}

\begin{figure}[h]
	\centering
	\includegraphics[width=\textwidth]{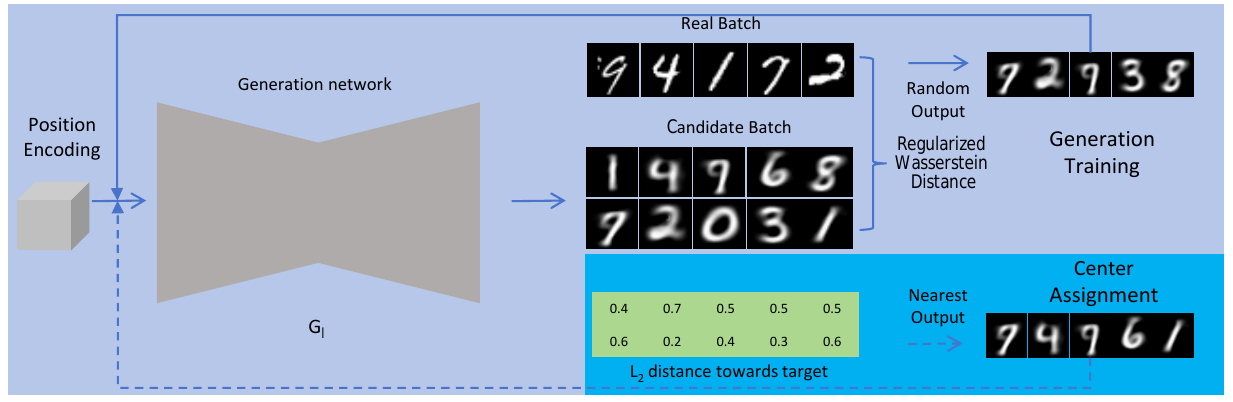}
\caption{Illustration of the proposed Discrete Generative Clustering (DGC) framework. The purple components denote the generative modeling module, while the blue components correspond to the clustering module.}
\label{framework}
\end{figure}

As illustrated in Figure~\ref{framework}, our model consists of $L$ sequentially connected sub-networks $\{{G}_1, {G}_2, \dots, {G}_L\}$.  
The first sub-network ${G}_1$ takes a fixed position embedding $z$ as input.
For $l = 2, \dots, L$, the $l$-th sub-network ${G}_l$ receives the output of the previous sub-network $y_{l-1}$ and generates a set of $B$ candidate outputs:
\begin{equation} 
\textbf{c}_l=\{c_{1}, c_{2}, \dots, c_{B}\} = {G}_l(y_{l-1}),
\end{equation} 
Every output of the model is selected from the candidates of the sub-network ${G}_l$ according to a stochastic hard selection rule described below:

\begin{equation}
	y_{{l}} \sim \text{Uniform}\{c_1,c_2,\dots,c_B\}.
\end{equation}

This stochastic selection confers generative diversity upon the model while preserving the advantages of the multi-candidate architecture learned during training. Empirically, we observe that random sampling maintains the quality of individual candidates and introduces beneficial variability, thereby mitigating the mode collapse often seen in deterministic and greedy selection schemes. 

\paragraph{Objective}
During training, we aim to align the distribution of generated outputs with the target data distribution.  
Given a mini-batch target images $\textbf{x}=\{x_1,\dots,x_n\}$ and a set of candidates $\textbf{c}_l=\{c_1,\dots,c_{n\times B}\}$ produced by the sub-network ${G}_l$, the training objective is the regularized Wasserstein distance between these two empirical measures.  The regularized optimal transport cost at $l$ layer is:
\begin{equation}
\theta^* \;=\; \arg\min_{\theta} \; W\bigl(\mathbf{x}, \mathbf{c}_l).
\end{equation}

The overall training objective aggregates the losses across all $L$ stages:
\begin{equation}
\mathcal{L}_{\text{total}}(\theta) = \frac{1}{L}\sum_{l=1}^{L}
W\bigl(\mathbf{x}, {\mathbf{c}}_l),
\end{equation}
where  $\theta$ denotes the trainable parameters of the entire network.  
This multi-level objective ensures that every sub-network learns to produce progressively refined representations, while the final stage aligns the generated distribution with the target distribution at the most fine-grained level.
Unlike the original DDN, DGC avoids two drawbacks: (i) its learning objective keeps every candidate involved in optimization, preventing candidates from becoming dead and ensuring all branches remain well-trained; (ii) it replaces hard nearest-neighbor assignment with stochastic candidate selection, thereby avoiding the density shift problem during generation. This issue is analogous to the bias caused by sequential softmax sampling in auto-regressive models.

\subsection{Unsupervised Clustering Analysis}
When the number of candidates $K = \prod_{l=1}^{L} B_l$ is set to a small value, the proposed generative model naturally performs unsupervised clustering. The intuition is that the generative modeling objective compresses the semantic information of the data distribution into a limited set of $K$ representative prototypes. These prototypes serve as cluster centroids in the data space.

In the clustering setting, real target images are available for matching. For each stage $l=1,\dots,L$ and each target sample $x_i$, we first select from the candidate set $\{c_{1},\dots,c_{B_l}\}$ produced by $G_l$ the candidate closest to $x_i$:
\begin{equation}
y_{i,l} = \arg\min_{b \in \{1,\dots,B_l\}} \left\| c_b - x_i \right\|_2.
\end{equation}
This selection is repeated across the entire DDN tree, assigning each target image to a centroid at every stage; the corresponding index is denoted by $y_{i,l}$.

\section{Experiments}

In this section, we present the experimental setup, quantitative and qualitative results, ablation studies, and additional analyses to validate the effectiveness of our proposed method.

\subsection{Experimental Setup}

\paragraph{Datasets}
We evaluate the proposed method on four benchmark datasets: MNIST~\cite{lecun1998gradient}, Fashion-MNIST~\cite{xiao2017fashion}, CIFAR-10~\cite{krizhevsky2009learning}, and COIL20~\cite{nene1996columbia}. Table~\ref{tab:datasets} summarizes the number of samples, original image resolutions, and class numbers. To ensure a fair comparison, all images are resized to $32\times32$ pixels before being fed into the network, except for COIL20, which is resized to $128\times128$ pixels. Unless otherwise stated, no data augmentation is applied, and raw pixel values are adopted as input features.

\begin{table}[!t]
	\centering
	\caption{Summary of datasets used in experiments. Original size refers to the native resolution.}
	\label{tab:datasets}
	\small
	\setlength{\tabcolsep}{8pt}
	\renewcommand{\arraystretch}{1.2}
	\begin{tabular}{lccc}
		\toprule
		Dataset & Samples & Original size & Classes \\
		\midrule
		MNIST          & 70,000 & $28\times28$   & 10 \\
		Fashion-MNIST  & 70,000 & $28\times28$   & 10 \\
		CIFAR-10       & 60,000 & $32\times32$   & 10 \\
		COIL20         & 1,440  & $128\times128$ & 20 \\
		\bottomrule
	\end{tabular}
\end{table}

\paragraph{Implementation Details}
The architecture of our method is detailed in Section~\ref{sec:method}.
The generator is a cascade of $L$ stacked discrete distribution blocks.
Each DDN adopts a U-Net-style encoder--decoder with instance normalization and SiLU activations. For example,  we use a two layer DDN with $B=5,2$ on the MNIST dataset.
The first DDN receives an initial spatial embedding; each subsequent DDN takes the feature map produced by its predecessor, which is projected and fused into the first encoder block via a residual addition.
The decoder outputs a tensor of size $B \times C \times H \times W$, which is reshaped into $B$ candidate reconstructions.
Each candidate is passed through a shared output head consisting of a $1\times1$ convolution followed by a $\tanh$ activation.
During training, the candidates are compared with the target image under the regularized Wasserstein distance, and the whole model is optimized end-to-end with Adam~\cite{kingma2014adam} at a learning rate of $2\times10^{-3}$. 
Our implementation is publicly available on GitHub.
\subsection{Clustering Results}
We evaluate clustering performance using clustering accuracy (ACC) and normalized mutual information (NMI). ACC is obtained by aligning predicted cluster assignments with ground-truth labels via the Hungarian algorithm, while NMI quantifies the normalized mutual information shared between the two labelings.
We first compare DGC with classical methods, including K-means, Gaussian Mixture Models (GMM), Hierarchical Clustering (HC), DBSCAN, and Spectral Clustering (SC). Then  we compare DGC with recent generative-model-based methods, including VaDE\cite{DBLP:conf/ijcai/JiangZTTZ17}, ClusterGAN\cite{mukherjee2019clustergan},GAN-EM\cite{DBLP:conf/ijcai/ZhaoWX0X19}, VaGAN\cite{DBLP:journals/tnn/YangFB22} and ClusterDDPM\cite{yan2025clusterddpm}.

As shown in Table~\ref{performance}, DGC consistently outperforms all classical baselines on every dataset. On MNIST, DGC achieves 62.14\% ACC, a relative improvement of 11.4\% over the strongest baseline SC (55.80\%), and also attains the best NMI (0.622), surpassing SC by 0.104. On COIL20, DGC reaches 51.09\% ACC, exceeding SC by 0.84 points. The improvements on Fashion-MNIST and CIFAR-10 are consistent as well, where DGC attains the highest ACC and NMI among classical methods.

To further assess DGC against recent generative-model-based clustering methods, we apply DGC as a secondary clustering module on the feature representations produced by ClusterGAN, resulting in the variant DGC + ClusterGAN. All generative baselines are evaluated under the same protocol for fairness. In particular, we re-implemented ClusterDDPM based on their official code. The original paper reports approximately 97.67\% accuracy on MNIST, while our reproduced result is 94.73\% under the same setting. As shown in Table~\ref{performance}, DGC+ClusterGAN achieves the best ACC and NMI on MNIST (96.42\% and 0.925) and Fashion-MNIST (68.33\% and 0.652), and achieves the best NMI on COIL20 (0.879). These results demonstrate that DGC serves as an effective and general clustering module that can be integrated with different generative models, providing robust improvements over their original clustering performance.
\begin{table}[!t]
	\centering
	\caption{Clustering performance comparison.}
	\label{performance}
	\small                                  
	\setlength{\tabcolsep}{3pt}            
	\renewcommand{\arraystretch}{1.2}
	\resizebox{\linewidth}{!}{%
		\begin{tabular}{l *{8}{c}}
			\toprule[1.2pt]
			\multirow{2}{*}{Method}
			& \multicolumn{2}{c}{MNIST}
			& \multicolumn{2}{c}{F-MNIST}
			& \multicolumn{2}{c}{CIFAR-10}
			& \multicolumn{2}{c}{COIL20} \\
			\cmidrule(lr){2-3} \cmidrule(lr){4-5}
			\cmidrule(lr){6-7} \cmidrule(lr){8-9}
			& ACC & NMI & ACC & NMI & ACC & NMI & ACC & NMI \\
			\midrule
			\multicolumn{9}{l}{\textbf{Classical clustering vs. DGC}} \\
			K-means    & 53.24 & 0.499 & 51.08 & 0.392 & 23.60 & 0.102 & 47.33 & 0.480 \\
			GMM        & 53.72 & 0.501 & 52.34 & 0.409 & 21.47 & 0.113 & 48.16 & 0.477 \\
			HC         & 52.10 & 0.485 & 49.75 & 0.378 & 20.12 & 0.095 & 46.20 & 0.465 \\
			DBSCAN     & 48.33 & 0.452 & 45.60 & 0.341 & 18.45 & 0.088 & 42.11 & 0.430 \\
			SC         & 55.80 & 0.518 & 53.90 & 0.421 & 25.30 & 0.121 & 50.25 & 0.492 \\
			\rowcolor{blue!6}
			DGC        & \textbf{62.14} & \textbf{0.627} & \textbf{54.81} & \textbf{0.428} & \textbf{25.38} & \textbf{0.138} & \textbf{51.09} & \textbf{0.511} \\
			\multicolumn{9}{l}{\textbf{Generative-model-based clustering vs. DGC}} \\
			VaDE       & 94.46 & 0.891 & 62.87 & 0.611 & 26.65 & 0.151 & 75.48 & 0.779 \\
			ClusterGAN & 95.00 & 0.890 & 63.00 & 0.640 & 28.12 & 0.156 & 77.00 & 0.811 \\
			GAN-EM     & 94.93 & 0.899 & 63.57 & 0.642 & 23.14 & 0.110 & 74.03 & 0.833 \\
			VaGAN  & 95.75 & 0.901 & 66.95 & 0.671 & 30.20 & 0.172 & 81.67 & 0.868 \\
			ClusterDDPM& 94.73 & 0.912 & 67.50 & 0.642 & 28.90 & 0.160 & 78.80 & 0.850 \\
			\rowcolor{blue!6}
			DGC+ClusterGAN & \textbf{96.42} & \textbf{0.925} & \textbf{68.33} & \textbf{0.652} & \textbf{29.15} & \textbf{0.165} & \textbf{80.05} & \textbf{0.879} \\
			\bottomrule
		\end{tabular}
	}%
\end{table}

\subsection{Qualitative Analysis}

\begin{figure}[!t]
	\centering

		\includegraphics[width=\linewidth]{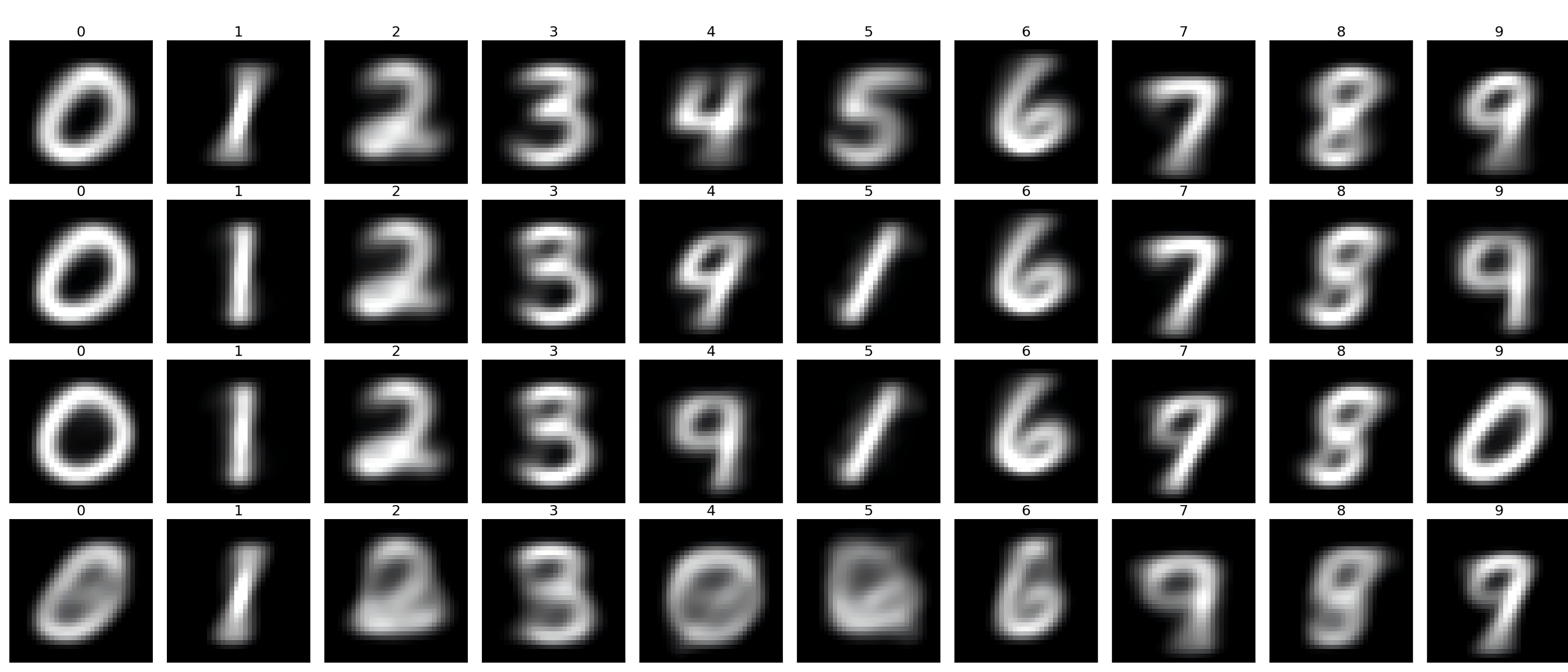}

	\caption{Comparison of cluster centers. Prototypes: Class mean, DDN, K-Means and GMM (columns = matched digit 0--9).
	}
	\label{fig:centers}
\end{figure}

\paragraph{Cluster Center Visualization}
Unlike traditional methods that produce abstract centroid parameters, our cluster centers are actual generated samples from the data distribution. Figure~\ref{fig:centers} compares the cluster centers produced by different algorithms. Notably, we do not apply any data normalization; the raw pixel values are used directly. Our method generates more meaningful and visually interpretable centers.

To verify whether the prototypes produced by DDN are statistically representative,
we compare them with the real class-wise means on MNIST.
As reported in Table~\ref{tab:prototype_quality},
DDN achieves the best average cosine similarity (0.9844), the lowest MSE (0.0233),
and the highest PSNR (22.35 dB) among all prototype estimators.
Moreover, DDN attains the best per-digit cosine similarity on 5 of 10 digits
and the best per-digit MSE on 5 of 10 digits.
In contrast, K-Means suffers from a catastrophic outlier on digit 9
(cosine similarity of 0.853 and MSE of 0.2061), and GMM degrades notably on digits 4--5.
These results indicate that generative prototypes are not only closer to the true class means
on average but also more stable across categories, making them more reliable for
nearest-prototype assignment in clustering.

\begin{table}[!t]
	\centering
	\caption{Prototype quality comparison on MNIST. All prototypes are compared with
		the real class-wise means over 60,000 training samples resized to $32\times32$.
		PSNR is computed with a peak value of 2 under $[-1,1]$ normalization.
		Cos.\ wins / MSE wins report the number of digits (out of 10) on which each method
		achieves the best per-digit cosine similarity / MSE.}
	\label{tab:prototype_quality}
	\small
	\renewcommand{\arraystretch}{1.2}
	\resizebox{\linewidth}{!}{%
		\begin{tabular}{lccccc}
			\toprule
			Method & Cos.$\uparrow$ & MSE$\downarrow$ & PSNR (dB)$\uparrow$ & Cos.\ wins & MSE wins \\
			\midrule
			K-Means & 0.9695 & 0.0438 & 19.60 & 3 & 2 \\
			GMM     & 0.9743 & 0.0349 & 20.59 & 2 & 3 \\
			\rowcolor{blue!6}
			\textbf{DDN} & \textbf{0.9844} & \textbf{0.0233} & \textbf{22.35} & \textbf{5} & \textbf{5} \\
 			\bottomrule
		\end{tabular}
	}%
\end{table}

\begin{figure*}[!t]
	\centering
	\begin{minipage}{0.22\textwidth}
		\includegraphics[width=\textwidth]{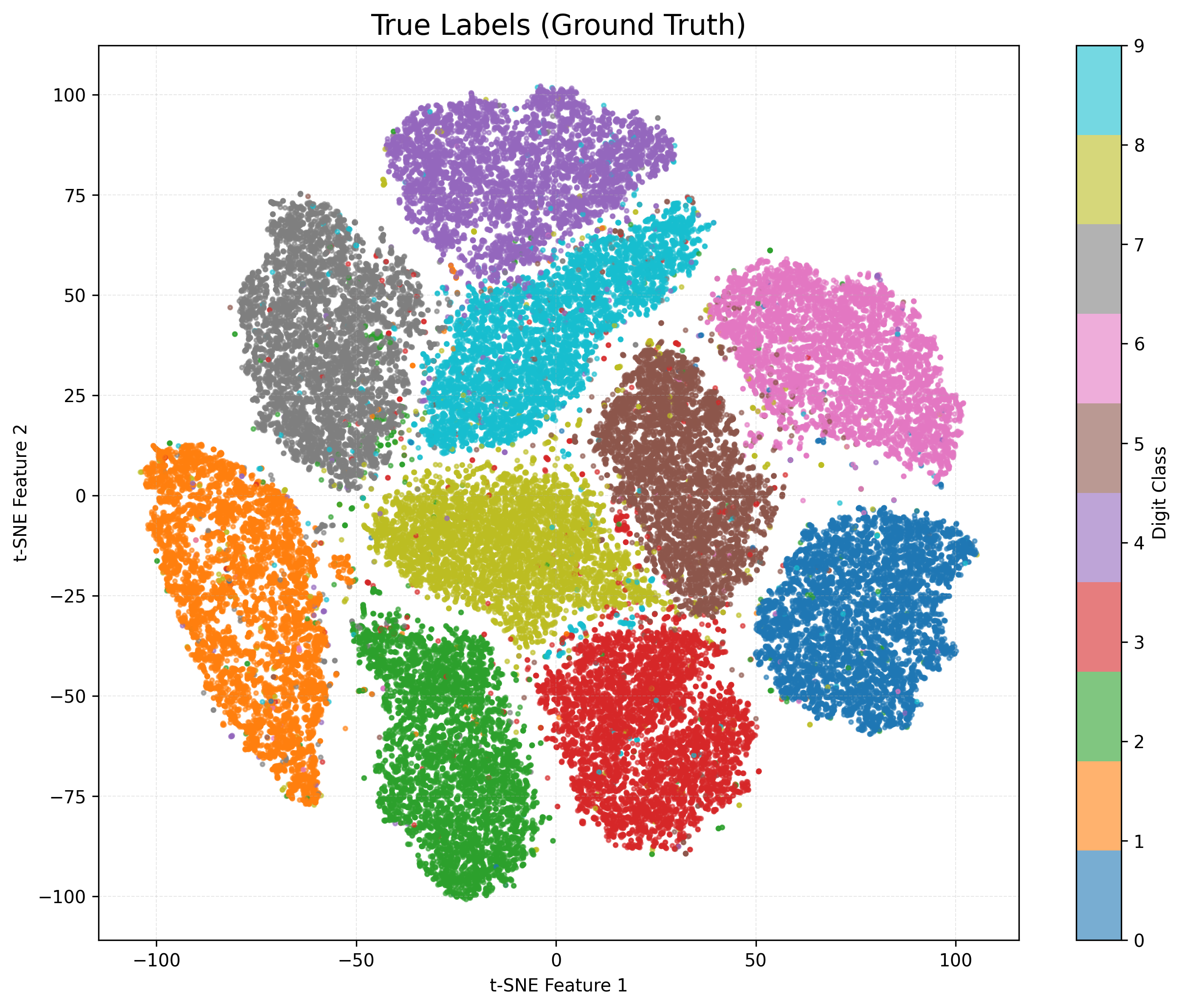}
		\caption*{(a) Ground truth}
	\end{minipage}
	\begin{minipage}{0.22\textwidth}
		\includegraphics[width=\textwidth]{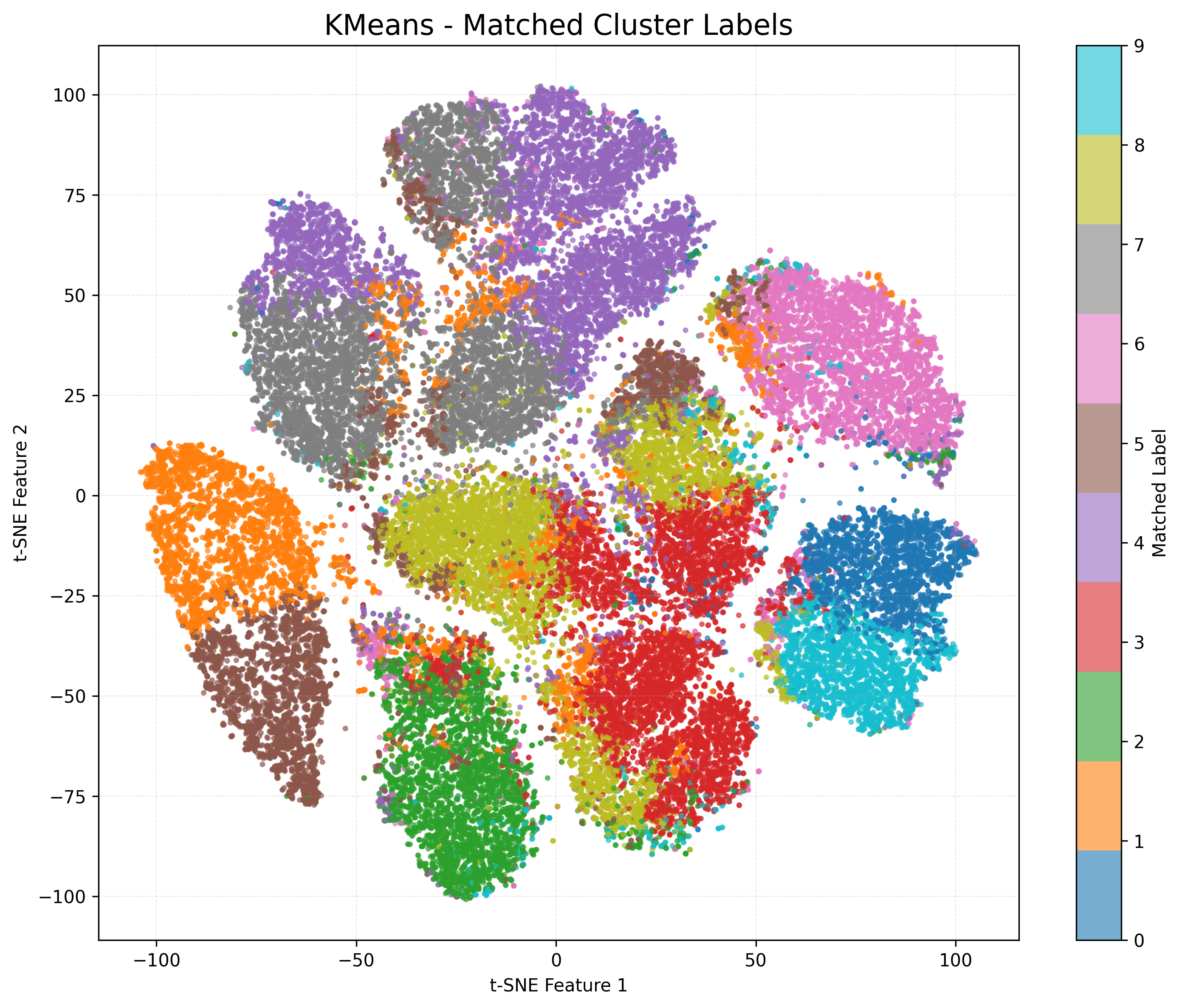}
		\caption*{(b) K-means}
	\end{minipage}
	\begin{minipage}{0.22\textwidth}
		\includegraphics[width=\textwidth]{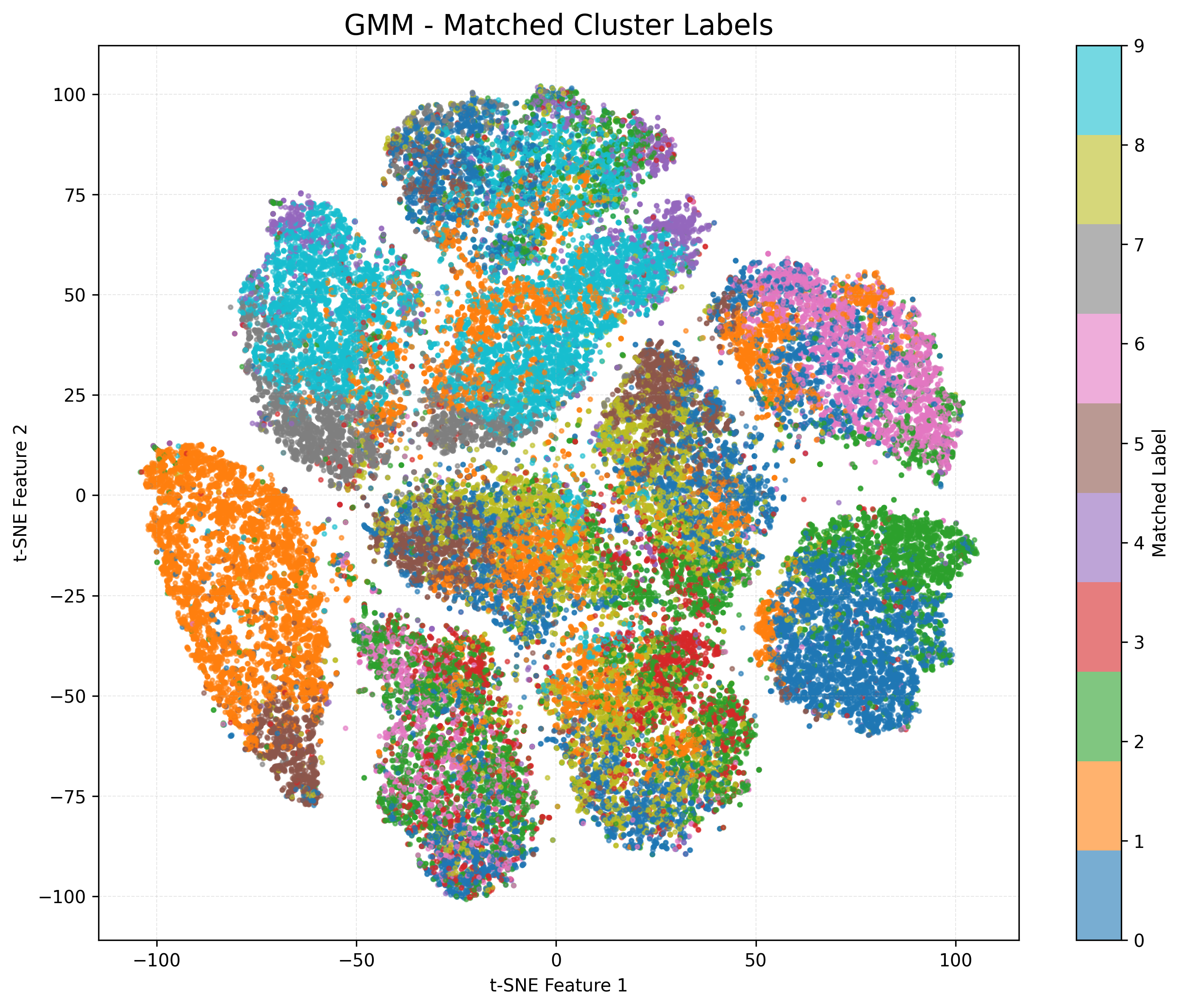}
		\caption*{(c) GMM}
	\end{minipage}
	\begin{minipage}{0.22\textwidth}
		\includegraphics[width=\textwidth]{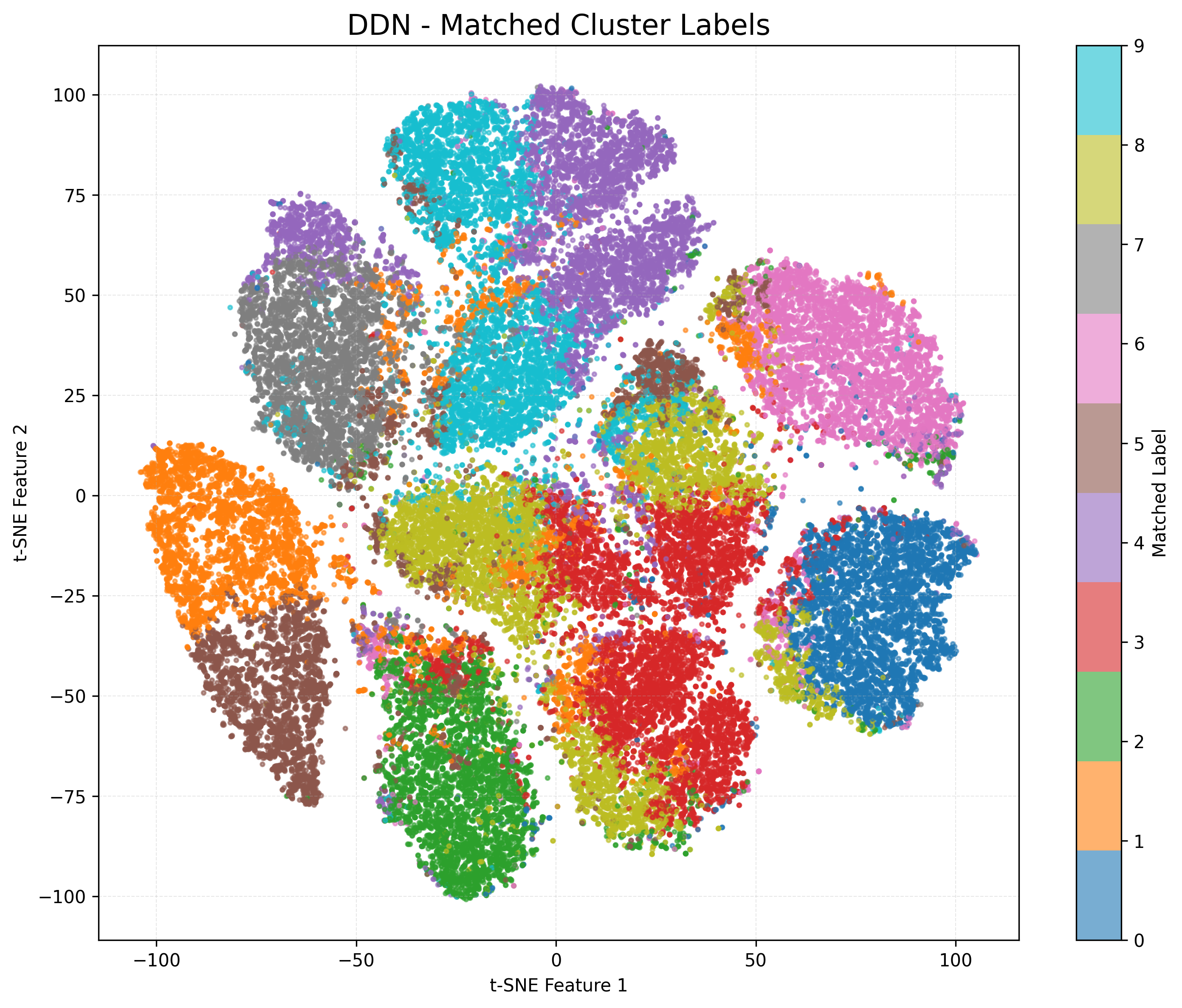}
		\caption*{(d) Ours}
	\end{minipage}
	\caption{t-SNE visualization on MNIST. Colors indicate ground-truth digit classes. Our method yields more separated clusters.}
	\label{fig:tsne}
\end{figure*}

\paragraph{t-SNE Visualization}
Figure~\ref{fig:tsne} visualizes the embedding spaces of different methods on MNIST. 
The t-SNE visualization on MNIST compares the embedding spaces of different clustering methods. The ground-truth labels form clear and well-separated clusters. In contrast, K-means and GMM produce heavily overlapping clusters with ambiguous boundaries, especially between similar digits. Our method yields compact, distinctly separated clusters that closely match the ground-truth structure, demonstrating that it learns more discriminative feature representations and preserves the data manifold better than traditional baselines.

\begin{figure}[!t]
\centering
\begin{subfigure}{0.43\textwidth}
	\centering
	\includegraphics[height=4.5cm]{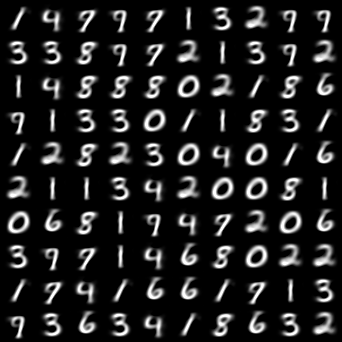}
	\caption{Continuous (MNIST)}
	\label{fig:sub1}
\end{subfigure}
\hfill
\begin{subfigure}{0.43\textwidth}
	\centering
	\includegraphics[height=4.5cm]{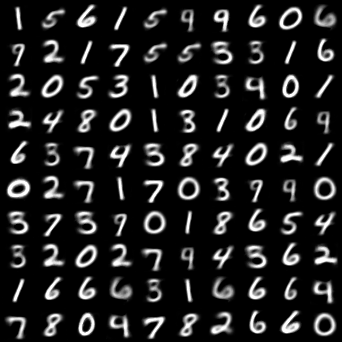}
	\caption{Discrete (MNIST)}
	\label{fig:sub2}
\end{subfigure}
\vskip 0.2cm
\begin{subfigure}{0.43\textwidth}
	\centering
	\includegraphics[height=4.5cm]{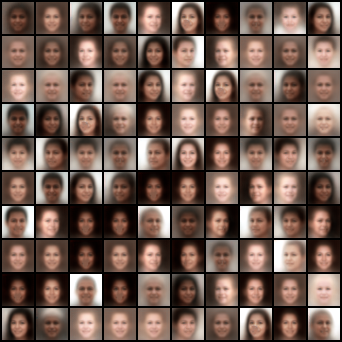}
	\caption{Continuous (FFHQ)}
	\label{fig:sub3}
\end{subfigure}
\hfill
\begin{subfigure}{0.43\textwidth}
	\centering
	\includegraphics[height=4.5cm]{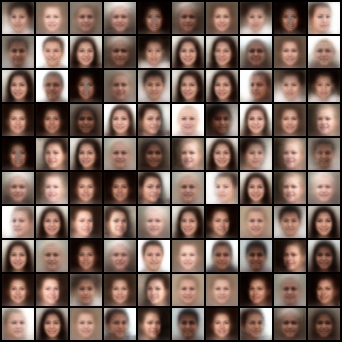}
	\caption{Discrete (FFHQ)}
	\label{fig:sub4}
\end{subfigure}
\caption{Generated samples. Continuous results are obtained by interpolation; discrete results are randomly selected candidates.}
\label{fig:generation}
\end{figure}

\paragraph{Generation Results}
Our model generates diverse samples by randomly selecting among the $K$ candidates. Figure~\ref{fig:generation} shows continuous and discrete generation results on MNIST and FFHQ. The discrete outputs maintain high visual quality while exhibiting diversity, which demonstrates the generation ability of DDN with the regularized Wasserstein distance.

\begin{figure*}[!t]
\centering
\begin{minipage}{0.19\textwidth}
	\centering
	\includegraphics[width=\textwidth]{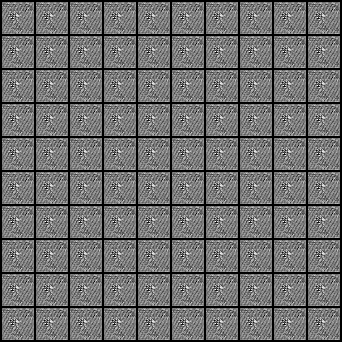}
	\caption*{(a) $K=1$  }
\end{minipage}
\begin{minipage}{0.19\textwidth}
	\centering
	\includegraphics[width=\textwidth]{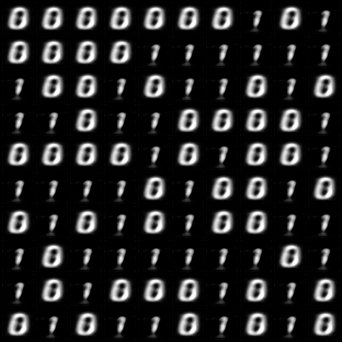}
	\caption*{(b) $K=2$}
\end{minipage}
\begin{minipage}{0.19\textwidth}
	\centering
	\includegraphics[width=\textwidth]{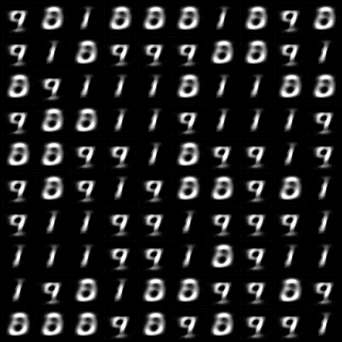}
	\caption*{(c) $K=3$}
\end{minipage}
\begin{minipage}{0.19\textwidth}
	\centering
	\includegraphics[width=\textwidth]{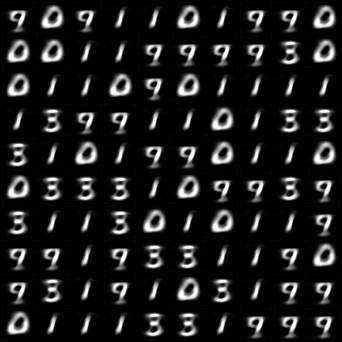}
	\caption*{(d) $K=4$}
\end{minipage}
\begin{minipage}{0.19\textwidth}
	\centering
	\includegraphics[width=\textwidth]{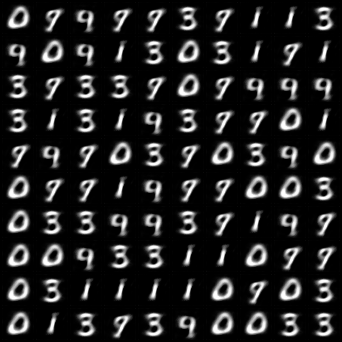}
	\caption*{(e) $K=5$}
\end{minipage}

\begin{minipage}{0.19\textwidth}
	\centering
	\includegraphics[width=\textwidth]{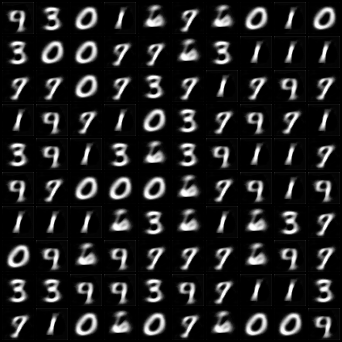}
	\caption*{(f) $K=6$}
\end{minipage}
\begin{minipage}{0.19\textwidth}
	\centering
	\includegraphics[width=\textwidth]{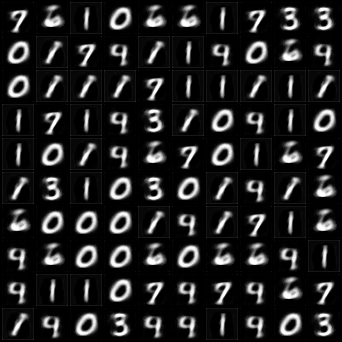}
	\caption*{(g) $K=7$}
\end{minipage}
\begin{minipage}{0.19\textwidth}
	\centering
	\includegraphics[width=\textwidth]{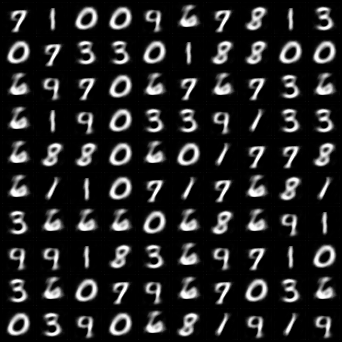}
	\caption*{(h) $K=8$}
\end{minipage}
\begin{minipage}{0.19\textwidth}
	\centering
	\includegraphics[width=\textwidth]{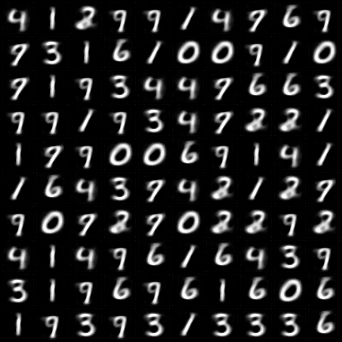}
	\caption*{(i) $K=9$}
\end{minipage}
\begin{minipage}{0.19\textwidth}
	\centering
	\includegraphics[width=\textwidth]{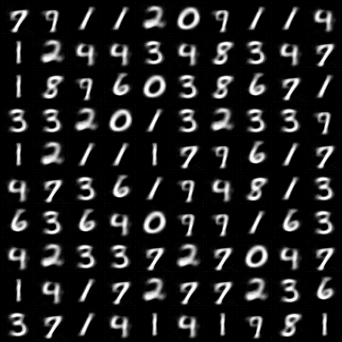}
	\caption*{(j) $K=10$}
\end{minipage}

\begin{minipage}{0.19\textwidth}
	\centering
	\includegraphics[width=\textwidth]{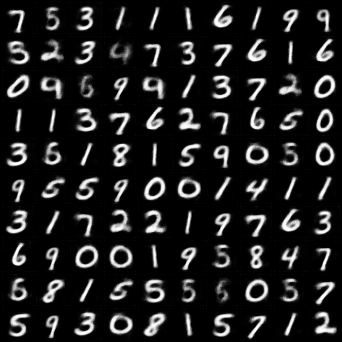}
	\caption*{(k) $K=100$}
\end{minipage}
\begin{minipage}{0.19\textwidth}
	\centering
	\includegraphics[width=\textwidth]{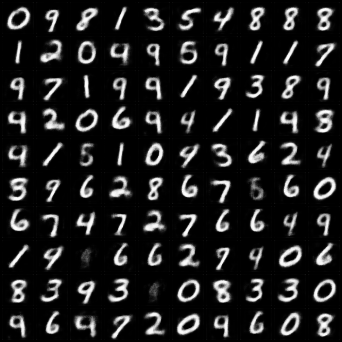}
	\caption*{(l) $K=1000$}
\end{minipage}
\begin{minipage}{0.19\textwidth}
	\centering
	\includegraphics[width=\textwidth]{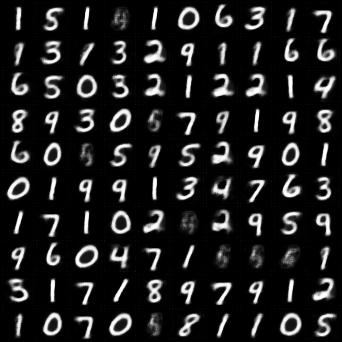}
	\caption*{(m) $K=10^4$}
\end{minipage}
\begin{minipage}{0.19\textwidth}
	\centering
	\includegraphics[width=\textwidth]{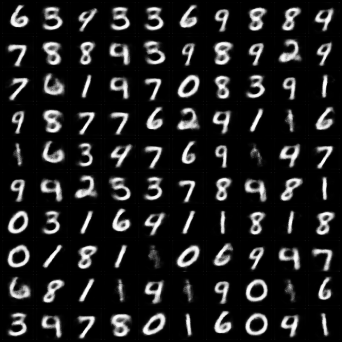}
	\caption*{(n) $K=10^5$}
\end{minipage}
\begin{minipage}{0.19\textwidth}
	\centering
	\includegraphics[width=\textwidth]{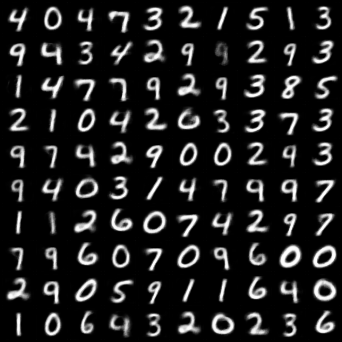}
	\caption*{(o) $K=10^6$}
\end{minipage}

\caption{Visualization of generation results with different candidate numbers $K$ on the MNIST dataset. When $K\le 10$, we use a single-layer DDN with the corresponding $K$. When $K>10$, we use $\log K$ DDN layers.}
\label{fig:k_aba}
\end{figure*}

\paragraph{Effect of Candidate Number $K$}
We investigate the impact of the number of candidate prototypes $K$ on the generative quality. As visualized in Fig.~\ref{fig:k_aba}, directly increasing $K$ within a small range (e.g., $K \le 10$) leads to noticeable improvements in the clarity and diversity of the generated digit images. However, naively enlarging $K$ beyond 10 introduces significant computational overhead. To address this, we adopt a tree-structured multi-layer DDN to organize the candidates, which effectively scales up $K$ (up to $10^6$) while strictly guaranteeing the clustering efficiency of $O(\log K)$. With this hierarchical design, larger $K$ values provide a richer prototype hypothesis space, enabling the model to capture finer data distributions and generate much clearer, more realistic, and separated digit prototypes compared to the baseline methods.

\begin{figure}[!t]
\centering
\includegraphics[width=\textwidth]{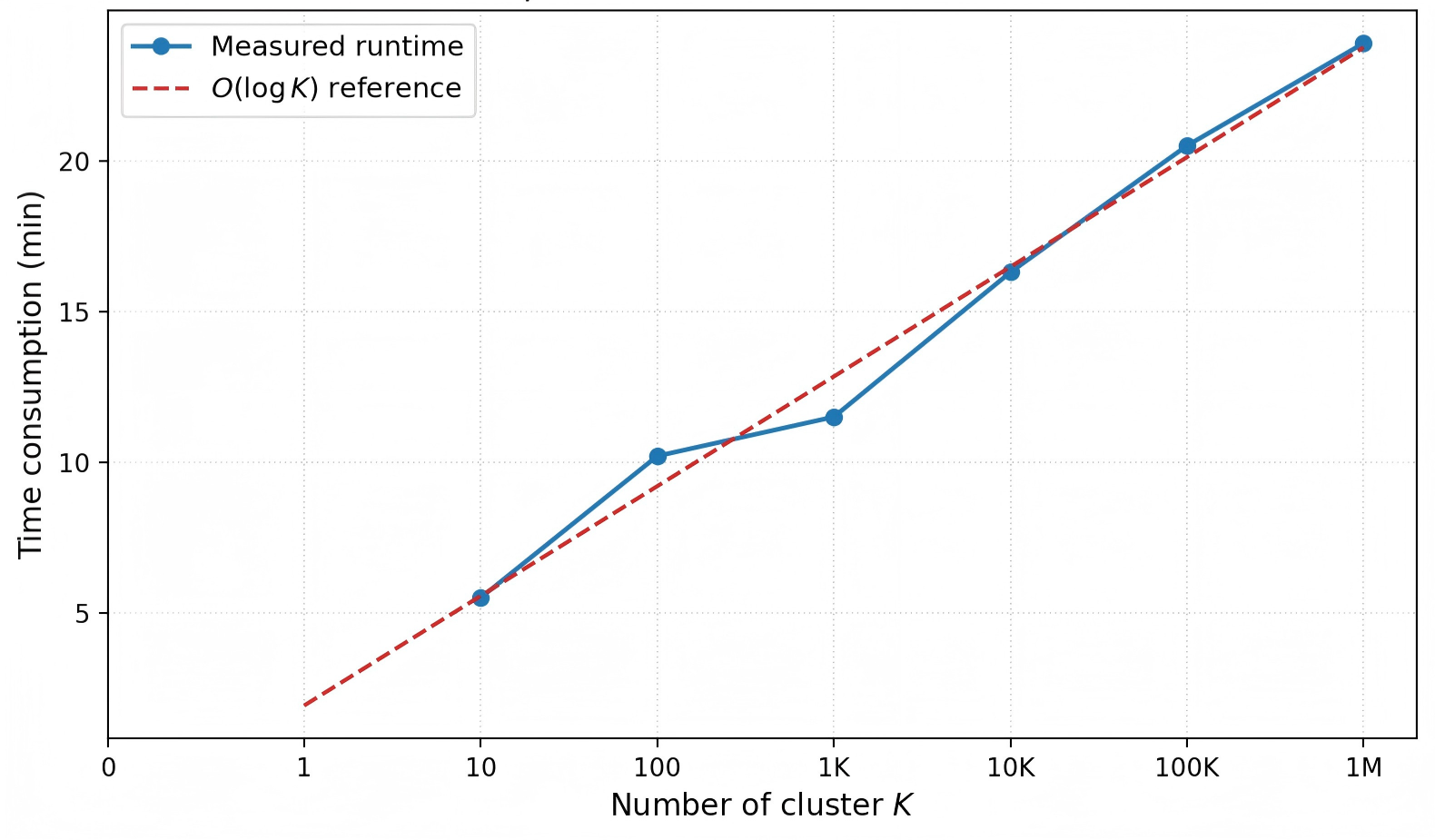}
\caption{Time consumption of different number of cluster K on the MNIST dataset.}
\label{fig:k_consumption}
\end{figure}

As shown in Fig.~\ref{fig:k_consumption}, we evaluate the runtime efficiency with respect to the number of candidates $K$ on the MNIST dataset. The results demonstrate that for small values of $K$ ($K \le 10$), the runtime remains nearly constant at approximately 5.5 min, which reflects the fixed baseline overhead of the process. 

As $K$ increases to $10^2$ and beyond, the runtime begins to rise. Notably, when plotted against $K$ on a logarithmic scale, the measured runtime curve aligns almost perfectly with the theoretical $O(\log K)$ reference line. This strong agreement visually verifies that our method achieves an $O(\log K)$ time complexity for large candidate sets. Even at $K=10^6$, the total time consumption is only 23.9 min, further confirming the high scalability and efficiency of the proposed approach.

\subsection{Ablation Studies}

	\paragraph{Effect of Network Width}
	We further examine the sensitivity of DGC to the network width $n_f$, i.e., the number of channels in the convolutional layers. Specifically, we vary $n_f \in \{4, 8, 16, 32\}$ while keeping all other hyperparameters fixed. As shown in Table~\ref{tab:ablation}, the clustering accuracy (ACC) remains largely stable across different widths: increasing $n_f$ from $4$ to $32$ improves the ACC by only $1.34$ percentage points on MNIST, and by less than $0.6$ points on Fashion-MNIST, CIFAR-10, and COIL20. Notably, even the compact network with $n_f=4$ already achieves competitive clustering performance, suggesting that the clustering objective relies primarily on the discriminative information captured by the learned prototypes rather than on the capacity of the network. In contrast, the visual quality of the generated samples improves noticeably with larger $n_f$: a wider network provides a richer hypothesis space and models finer details of the data distribution, yielding clearer and more realistic prototypes. We therefore adopt $n_f = 32$ as the default configuration to balance clustering performance and generative fidelity.
	
\begin{table}[h]
	\centering
	\caption{Ablation study on network width $n_f$, Sinkhorn–Knopp loss presence, and weight $\lambda$.}
	\label{tab:ablation}
	\small
	\setlength{\tabcolsep}{3pt}
	\renewcommand{\arraystretch}{1.2}
	\resizebox{\linewidth}{!}{%
		\begin{tabular}{lcccccccc}
			\toprule
			\multirow{2}{*}{Config}
			& \multicolumn{2}{c}{MNIST}
			& \multicolumn{2}{c}{F-MNIST}
			& \multicolumn{2}{c}{CIFAR-10}
			& \multicolumn{2}{c}{COIL20} \\
			\cmidrule(lr){2-3} \cmidrule(lr){4-5}
			\cmidrule(lr){6-7} \cmidrule(lr){8-9}
			& {ACC} & {NMI} & {ACC} & {NMI} & {ACC} & {NMI} & {ACC} & {NMI} \\
			\midrule
			
			\multicolumn{9}{@{}l}{\textbf{(a) Effect of network width $n_f$ (with $\lambda=0.1$)}} \\
			4  & 60.80 & 0.612 & 54.30 & 0.421 & 25.00 & 0.132 & 50.60 & 0.496 \\
			8  & 61.00 & 0.621 & 54.60 & 0.430 & 25.10 & 0.134 & 50.87& 0.503 \\
			16 & 61.20 & 0.624 & 54.70 & 0.427 & 25.30 & 0.137 & 51.00 & 0.508 \\
			\rowcolor{blue!6}
			32 & \textbf{62.14} & \textbf{0.627} & \textbf{54.81} & \textbf{0.428} & \textbf{25.38} & \textbf{0.138} & \textbf{51.09} & \textbf{0.511} \\
			
			\rowcolor{gray!15}
			\multicolumn{9}{@{}l}{\textbf{(b) DDN baseline with SPLIT-AND-PRUNE}} \\
			32 & 54.20 & 0.542 & 47.80 & 0.371 & 22.20 & 0.122 & 44.40 & 0.454 \\
			
			\rowcolor{gray!15}
			\multicolumn{9}{@{}l}{\textbf{(c) DDN baseline without Sinkhorn–Knopp loss}} \\
			32 & 11.50 & 0.051 & 14.20 & 0.092 & 10.80 & 0.021 & 6.50 & 0.031 \\
			
			\rowcolor{gray!15}
			\multicolumn{9}{@{}l}{\textbf{(d) Sensitivity to Sinkhorn–Knopp loss weight $\lambda$ (at $n_f=32$)}} \\
			0.01 & 55.20 & 0.453 & 49.30 & 0.342 & 21.40 & 0.091 & 45.10 & 0.407 \\
			0.05 & 59.80 & 0.558 & 52.60 & 0.394 & 23.90 & 0.125 & 49.20 & 0.478 \\
			\rowcolor{blue!6}
			0.10 & \textbf{62.14} & \textbf{0.627} & \textbf{54.81} & \textbf{0.430} & \textbf{25.38} & \textbf{0.140} & \textbf{51.09} & \textbf{0.511} \\
			0.50 & 36.90 & 0.282 & 32.40 & 0.223 & 13.10 & 0.042 & 29.80 & 0.292 \\
			\bottomrule
		\end{tabular}
		
	}%
\end{table}
\paragraph{Effect of regularized Wasserstein distance}We further ablate the effect of Sinkhorn–Knopp loss in Table~\ref{tab:ablation}. \textbf{DDN baseline with SPLIT-AND-PRUNE} represents the original DDN's results and \textbf{DDN baseline without Sinkhorn–Knopp loss} removes SPLIT-AND-PRUNE operation and Chain Dropout from DNN.

 First, we compare DGC with the original DDN baseline with SPLIT-AND-PRUNE (row (b)). DGC consistently outperforms this baseline, e.g., 62.14\% vs.\ 54.20\% ACC on MNIST, 54.81\% vs.\ 47.80\% on Fashion-MNIST, 25.38\% vs.\ 22.20\% on CIFAR-10, and 51.09\% vs.\ 44.40\% on COIL20, confirming the benefit of our Sinkhorn-based objective and the hierarchical candidate selection. Removing the Wasserstein distance term (row (c)) causes a catastrophic degradation across all datasets. On MNIST, the ACC drops from 62.14\% to 11.50\%, which is essentially random guessing, and similar collapses are observed on Fashion-MNIST (14.20\%), CIFAR-10 (10.80\%), and COIL20 (6.50\%, whose random baseline is 5\%). This confirms that without an optimal-transport-based geometric constraint, the generated prototypes fail to align with the true data distribution and clustering degenerates into trivial solutions.

The weight $\lambda$ of the regularized Wasserstein distance plays a critical role. At $\lambda=0.01$, the performance is markedly lower (e.g., 55.20\% ACC on MNIST). This is because a small $\lambda$ weakens the entropic regularization and makes the Sinkhorn–Knopp iteration converge very slowly; with the iteration number truncated at 30, the approximation error becomes non-negligible, and the computed regularized Wasserstein distance no longer reflects the true transport cost. As $\lambda$ increases to 0.05, the accuracy improves substantially (59.80\% on MNIST), and the best results are obtained at $\lambda=0.10$ (62.14\% on MNIST), where the regularized Wasserstein distance is accurately approximated within 30 iterations while still preserving sufficient geometric structure. However, when $\lambda$ is too large, e.g., $\lambda=0.50$, the entropy regularization dominates the transport cost and the regularized Wasserstein distance gradually deviates from the true earth mover's distance. Consequently, the geometric information is no longer faithfully preserved, and the ACC drops sharply to 36.90\% on MNIST, with analogous declines on the other datasets. These results demonstrate that $\lambda$ should be chosen in a moderate range to balance the approximation accuracy of the Sinkhorn iteration and the geometric fidelity of the optimal transport objective.

\begin{figure}[!t]
\centering
\begin{subfigure}{0.43\textwidth}
	\centering
	\includegraphics[height=4.5cm]{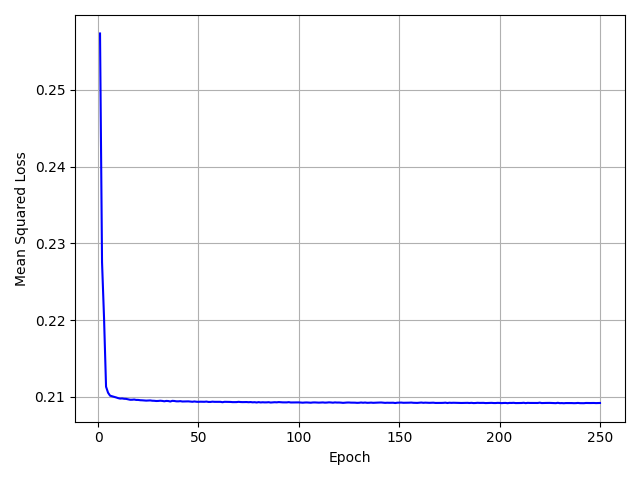}
	\caption{ Mean squared loss}
	\label{fig:csub1}
\end{subfigure}
\hfill
\begin{subfigure}{0.43\textwidth}
	\centering
	\includegraphics[height=4.5cm]{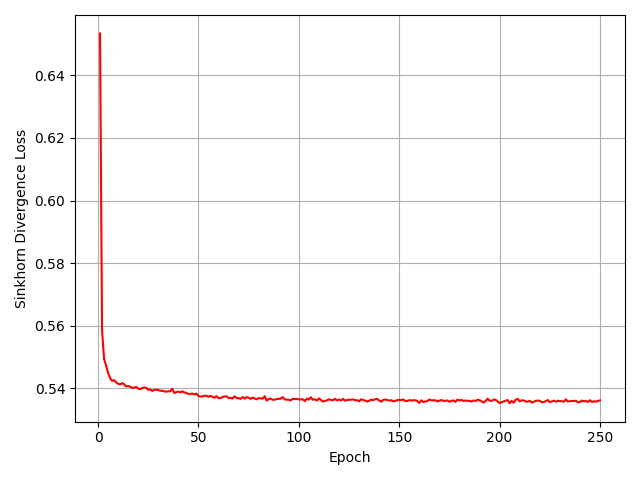}
	\caption{Sinkhorn–Knopp loss}
	\label{fig:csub2}
\end{subfigure}

\caption{Training loss comparison of mean squared loss and Sinkhorn–Knopp loss.}
\label{fig:convergence}
\end{figure}

\paragraph{Convergence Analysis}
Figure~\ref{fig:convergence} compares the training loss curves of DDN's  $\ell_2$ loss and the proposed regularized Wasserstein distance. In contrast to adversarial training, which often suffers from instability and oscillations, our method converges smoothly without severe fluctuations. This desirable behavior can be attributed to the strict convexity of the entropy-regularized optimal transport problem with respect to the transport plan, which yields well-behaved gradients throughout training. The  $\ell_2$ loss  prones to getting trapped in poor local optima, as reflected by a loss curve that plateaus early with little subsequent improvement. With the regularized Wasserstein distance, the loss decreases steadily and reaches a good solution, exhibiting only minor fluctuations typical of stochastic optimization.

\section{Related Work}

Our proposed DGC shares conceptual similarities with several classic clustering paradigms, yet differs fundamentally in both formulation and practical behavior.

\paragraph{K-means}

K-means seeks to partition data into \(K\) clusters by minimizing the within-cluster variance~\cite{lloyd1982least}. While intuitive, it is sensitive to initialization, requires a fixed \(K\), and operates on raw feature space without learning representations. DGC, in contrast, learns the cluster centers end-to-end via a deep neural network. Moreover, DGC optimizes a regularized Wasserstein distance objective that preserves global geometric structure~\cite{cuturi2013sinkhorn, genevay2018sinkhorn}, in contrast to the squared Euclidean distance used in K-means, whose non-convex objective provides no guarantee of reaching a global optimum.

\paragraph{Self-Organizing Maps}
SOM is a neural network-based clustering method that preserves topological structure by mapping high-dimensional data onto a low-dimensional grid~\cite{kohonen1990self}. Both SOM and DGC share a similar objective: they try to assign each data point to its nearest prototype. However, DGC offers several advantages. First, its tree-structured architecture enables efficient inference with \(O(\log K)\) complexity, whereas SOM typically requires scanning all prototypes (\(O(K)\)). Second, SOM focuses on preserving local topology (neighborhood relationships), while DGC aims to produce cluster centers that are the most representative samples from the data distribution, offering clearer interpretability. Third, DGC leverages a deep neural network to learn the underlying data structure automatically, avoiding hand-crafted rules for handling nearby node similarity.

\paragraph{Clustering with Generative Models}
A number of recent studies have sought to connect generative modeling with clustering. ClusterGAN \cite{mukherjee2019clustergan} and ClusterDDPM \cite{yan2025clusterddpm}, for instance, adopt a two-stage pipeline in which a generative model first learns unsupervised deep representations of the data, and a conventional clustering algorithm such as K-means or GMM is then applied to the extracted features. While effective in practice, these methods treat the generative model merely as a feature extractor and do not exploit the structural relationship between the generation process and the clustering objective. In a different line of work, optimal transport has been introduced into clustering as a distance measure \cite{chakraborty2020hierarchical,delbarrio2019robust,laclau2017coclustering}; however, these approaches employ Wasserstein distance purely as a metric for comparing clusters, rather than as a distribution fitting criterion. Consequently, existing methods generally regard generation and clustering as two independent tasks. The former aims to model the global data distribution for realistic sampling, while the latter seeks representative centers for data partitioning. In this paper, we argue that c lustering can be naturally formulated as a constrained generative modeling problem.
\bibliographystyle{elsarticle-num} 
\bibliography{iclr2025_conference}

\end{document}